\documentclass{article}

\usepackage{natbib}
\usepackage[final]{nips_2017}

\usepackage[utf8]{inputenc} % allow utf-8 input
\usepackage[T1]{fontenc}    % use 8-bit T1 fonts
\usepackage{hyperref}       % hyperlinks
\usepackage{url}            % simple URL typesetting
\usepackage{booktabs}       % professional-quality tables
\usepackage{amsfonts}       % blackboard math symbols
\usepackage{nicefrac}       % compact symbols for 1/2, etc.
\usepackage{microtype}      % microtypography
\usepackage{dsfont}
\usepackage{amssymb}
\usepackage{amsmath}
\usepackage[font=small,labelfont=bf]{caption}
\usepackage{multirow}
\usepackage{graphicx}
\usepackage{varwidth}
\usepackage[toc,page]{appendix}
\usepackage{makecell}
\usepackage{todonotes}
\usepackage{algorithm}% http://ctan.org/pkg/algorithms
\usepackage{algpseudocode}% http://ctan.org/pkg/algorithmicx

\newcommand{\lp}{\left(}
\newcommand{\rp}{\right)}
\newcommand{\lb}{\left[}
\newcommand{\rb}{\right]}

\newcommand{\E}{{\mathbb{E}}}

\newcommand{\wvec}{{\bf{w}}}

\newcommand{\mvec}{{\bf{m}}}

\newcommand{\phivec}{{\boldsymbol{\phi}}}
\newcommand{\muvec}{{\boldsymbol{\mu}}}
\newcommand{\pivec}{{\boldsymbol{\pi}}}

\newcommand{\lambdavec}{{\boldsymbol{\lambda}}}
\newcommand{\thetavec}{{\boldsymbol{\theta}}}

\newcommand{\psivec}{{\boldsymbol{\psi}}}

\title{Improved Bayesian Compression}

\author{
 Marco Federici\\
  University of Amsterdam\\
  \texttt{marco.federici@student.uva.nl}\And
  Karen Ullrich\\
  University of Amsterdam\\
\texttt{karen.ullrich@uva.nl}\And
  Max Welling\\
  University of Amsterdam\\
  Canadian Institute for Advanced Research (CIFAR) \\
\texttt{welling.max@gmail.com}
}

\begin{document}

\maketitle

\vspace{-0.8cm}\section{Variational Bayesian Networks for Compression}
% Establish a niche:
% Why is compression important
Compression of Neural Networks (NN) has become a highly studied topic in recent years. The main reason for this is the demand for industrial scale usage of NNs such as deploying them on mobile devices, storing them efficiently, transmitting them via band-limited channels and most importantly doing inference at scale. 
%
% While they seem to rely on millions of parameters for training several studies suggest that one can reduce them in size without significant loss of accuracy \citep{hinton,denil}.
% Name main approaches to compression
% A popular compression schemes first proposed by \cite{han}, relies on pruning unnecessary weights and quantizing the remaining ones. 
% ??????
%
% Close down your niche:
% Variational Bayesian Networks for compression
% The well established connection between variational learning and the minimum description principle had recently been revisited by various authors to directly tackle compression of neural networks. The argument goes as follows, the variational lower bound can directly be linked to model compression in particular the KL-divergence from model posterior to prior. Choosing a suitable model prior determines the success of the compression scheme. 
There have been two proposals that show strong results, both using empirical Bayesian priors: 
(i) \citet{2017arXiv170204008U} show impressive compression results by use of an adaptive Mixture of Gaussian prior on independent delta distributed weights. This idea has initially been proposed as \textit{Soft-Weight Sharing} by \citet{Nowlan:1992:SNN:148167.148169} but was never demonstrated to compress before.
(ii) Equivalently, \citet{2017arXiv170105369M} use \textit{Variational Dropout} \citep{2015arXiv150602557K} to prune out independent Gaussian posterior weights with high uncertainties. To achieve high pruning rates the authors
refined the originally proposed approximations to the KL-divergence and a different parametrization to increase the stability of the training procedure.
% The results of this technique have recently been improved by applying the same prior to groups of weights such as convolutional kernels or weight matrix rows. This allows for faster inference and compression simultaneously \citep{louizos2017bayesian}.
%
In this work, we propose to join these two somewhat orthogonal compression schemes since (ii) seems to prune out more weights but does not provide a technique for quantization such as (i). 
We find our method to outperform both of the above.

\section{Method}

% In this section we briefly describe the Bayesian Inference and Stochastic Variational frameworks and show their application in the context of models compression.

Given a dataset $\mathcal{D}=\left\{ x_i,y_i\right\}_{i=1}^N$ and a model parametrized by a weight vector $\wvec$, the learning goal consists in the maximization of the posterior probability of the parameters given the data $p\lp \wvec | \mathcal{D}\rp$. Since this quantity involves the computation of intractable integrals, the original objective is replaced with a lower bound obtained by introducing an approximate parametric posterior distribution $q_\phi\lp\wvec\rp$.
In the \textit{Variational Inference} framework, the objective function is expressed as a \textit{Variational Lower Bound}:
% \begin{align}
% \mathcal{L}\lp \phi \rp &= L_{\mathcal{D}}\lp\phi\rp - D_{KL}\lp q_{\phi}\lp \wvec\rp || p\lp\wvec\rp\rp
% \label{eq:elbo}
% \end{align}
% \begin{align*}
% \text{where}\ \ \ \ \ L_{\mathcal{D}}\lp\phi\rp &= \sum_{i=1}^N \E_{\wvec\sim q_{\phi}\lp\wvec\rp}\lb \log p\lp y_n| x_n, \wvec\rp\rb
% \end{align*}
\begin{align}
\mathcal{L}\lp \phi \rp &=\underbrace{\sum_{i=1}^N \E_{\wvec\sim q_{\phi}\lp\wvec\rp}\lb \log p\lp y_n| x_n, \wvec\rp\rb}_{L_{\mathcal{D}}\lp\phi\rp} - D_{KL}\lp q_{\phi}\lp \wvec\rp || p\lp\wvec\rp\rp
\label{eq:elbo}
\end{align}

%and it can be approximated using a re-parametrization trick and batch-based Monte Carlo simulation (CITE).On the other hand 
The first term $L_{\mathcal{D}}\lp\phi\rp$ of the equation represents the log-likelihood of the model predictions, while the second part $ D_{KL}\lp q_{\phi}\lp \wvec\rp || p\lp\wvec\rp\rp$ stands for the KL-Divergence between the weights approximate posterior $q_{\phi}\lp \wvec\rp $ and their prior distribution $p\lp \wvec\rp$. This term works as a regularizer whose effects on the training procedure and tractability depend on the chosen functional form for the two distributions.
In this work we propose the use of a joint distribution over the $D$-dimensional weight vector $\wvec$ and their corresponding centers $\mvec$.
\begin{table}[!ht]
\centering
\caption{Choice of probability distribution for the approximate posterior $q_{\boldsymbol{\phi}}\lp\wvec,\mvec\rp$ and the prior distribution $p_{\boldsymbol{\psi}}\lp \wvec,\mvec\rp$. $GM_{\psivec}\lp m_i \rp$ represent a Gaussian mixture model over $m_i$ parametrized with $\psivec$.\\}
\label{tab:distr}
\begin{tabular}{ll}
\hline
Distribution factorization                                                             & Functional forms                                            \\ \hline
$\displaystyle q_{\boldsymbol{\phi}}\lp\wvec,\mvec\rp=\prod_{i=1}^{D}\lp q_{\sigma_i}\lp w_i|m_i\rp q_{\theta_i}\lp m_i\rp\rp$& {$\!\begin{aligned}&q_{\sigma_i}\lp w_i|m_i\rp = \mathcal{N}\lp w_i|m_i,\sigma_i^2\rp\\&q_{\theta_i}\lp m_i\rp = \delta_{\theta_i}\lp m_i\rp\end{aligned}$}            \\\hline
$\displaystyle p_{\boldsymbol{\psi}}\lp \wvec,\mvec\rp=\prod_{i=1}^D \lp p\lp w_i\rp\ p_{\boldsymbol{\psi}}\lp m_i\rp\rp$       & {$\!\begin{aligned}&p\lp w_i\rp \propto 1/|w_i|\\&p_{\boldsymbol{\psi}}\lp m_i\rp = GM_{\psivec}\lp m_i\rp\end{aligned}$}
\end{tabular}
\end{table}

Table \ref{tab:distr} shows the factorization and functional form of the prior and posterior distributions.
% \begin{align*}
% q_{\boldsymbol{\phi}}\lp\wvec,\mvec\rp = \prod_{i=1}^{D}\lp q_{\sigma_i}\lp w_i|m_i\rp q_{\theta_i}\lp m_i\rp\rp
% \end{align*}
Each conditional posterior $q_{\sigma_i}\lp w_i|m_i\rp$ is represented with a Gaussian distribution with a variance $\sigma_i^2$ around a center $m_i$ defined as a delta peak determined by the parameter $\theta_i$.
% \begin{align*}
% q_{\sigma_i}\lp w_i|m_i\rp = \mathcal{N}\lp w_i|m_i,\sigma_i^2\rp,\ \ \ q_{\theta_i}\lp m_i\rp = \delta_{\theta_i}\lp m_i\rp
% \end{align*}

On the other hand, the joint prior is modeled as a product of independent distributions over $\wvec$ and $\mvec$.
% \begin{align*}
% p_{\boldsymbol{\psi}}\lp \wvec,\mvec\rp = \prod_{i=1}^D \lp p\lp w_i\rp\ p_{\boldsymbol{\psi}}\lp m_i\rp\rp
% \end{align*}
Each $p\lp w_i\rp$ represents a log-uniform distribution, while $p_{\boldsymbol{\psi}}\lp m_i\rp$ is a mixture of Gaussian distribution parametrized with $\boldsymbol{\psi}$ that represents the $K$ mixing proportions $\pivec$, the mean of each Gaussian component $\muvec$ and their respective precision $\lambdavec$.
% \begin{align*}
% p\lp w_i\rp \propto \frac{1}{|w_i|},\ \ \ p_{\boldsymbol{\psi}}\lp m_i\rp = \sum_{k=1}^K \pi_k\ \mathcal{N}\lp m_i| \mu_k, \lambda_k^{-1}\rp = GM\lp m_i |\boldsymbol{\psi}\rp
% \end{align*}
In this settings, the KL-Divergence between the prior and posterior distribution can be expressed as:
\begin{align}
D_{KL}\lp q_{\boldsymbol{\phi}}\lp\wvec,\mvec\rp || p_{\boldsymbol{\psi}}\lp \wvec,\mvec\rp\rp &=\sum_{i=1}^D\lp D_{KL}\lp \mathcal{N}\lp w_i|\theta_i,\sigma_i^2\rp||\frac{1}{\lvert w_i\rvert}\rp +\log GM_{\psivec}\lp m_i=\theta_i \rp\rp+C
\label{eq:klend}
\end{align}
Where $D_{KL}\lp \mathcal{N}\lp w_i|\theta_i,\sigma_i^2\rp||1/\lvert w_i\rvert\rp$ can be effectively approximated as described in \citet{2017arXiv170105369M}. A full derivation can be found in appendix \ref{app:kl}.

The zero-centered heavy-tailed prior distribution on $\wvec$ induces sparsity in the parameters vector, at the same time the adaptive mixture model applied on the weight centers $\mvec$ forces a clustering behavior while adjusting the parameters $\psivec$ to better match their distribution.
The final expression for the training objective is consequently represented by:
\begin{align*}
\mathcal{L}\lp \phivec,\psivec \rp &= L_{\mathcal{D}}\lp\phivec\rp - D_{KL}\lp q_{\phivec}\lp \wvec,\mvec\rp || p_{\psivec}\lp\wvec,\mvec\rp\rp\\
&=L_{\mathcal{D}}\lp\phi\rp  -\sum_{i=1}^D\lp D_{KL}\lp \mathcal{N}\lp w_i|\theta_i,\sigma_i^2\rp||\frac{1}{\lvert w_i\rvert}\rp +\log GM_{\psivec}\lp m_i=\theta_i \rp\rp
\end{align*}
Additional details regarding the optimization procedure and the parameter initialization can be found in appendix \ref{app:training}

\section{Model Compression}
\label{sec:compression}
\begin{figure}[!ht]
\caption{Graphical representation of the model compression pipeline. First (i) the model weight are clustered into predetermined number of components, then (ii) empty rows and column are removed. (iii) The weight matrices are then converted into the Compressed Sparse Row format and (iv) the floating point representation of the remaining weights is replaced with an optimal Huffman code.}
\label{fig:compression}
\includegraphics[width=\textwidth]{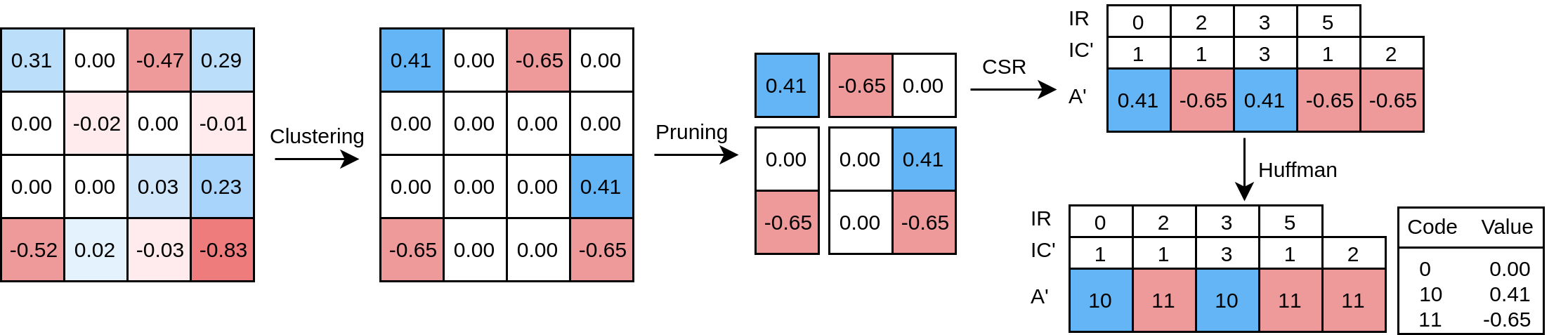}
\end{figure}

In this section, we briefly describe the post-training model processing, with a focus on the compression scheme. The procedure (that is based on the pipeline described in \citet{2015arXiv151000149H}) is schematically represented in Figure \ref{fig:compression} and consists in 4 steps:  (i) \textit{Weight Clustering}, (ii) \textit{Model Pruning}, (iii) \textit{Compressed Sparse Row Encoding} and (iv) \textit{Huffman Encoding}.

\paragraph{Weight Clustering}
At the end of the training procedure, the weight posterior means $\thetavec$ are clustered according to the mixture model defined by the parameters $\psivec$. Each parameter $\theta_i$ is collapsed into the mean $\mu_k$ of the mixture component $k$ that assigns it the highest probability. The resulting approximation doesn't compromise the model performances because the training objective maximizes the likelihood of $\thetavec$ adapting $\psivec$ and vice-versa. Note that one of the mixture component is fixed to $0$ to induce sparsity in the weight matrices. Additional details regarding this procedure can be found in \citet{2017arXiv170204008U}.
\paragraph{Model Pruning}
Empty rows and columns in the weight matrices can be removed from the model since they don't affect its prediction. If the $j$-th row of the weight matrix at layer $i$ is filled with zeros and has a zero bias, it can be removed together with the $j$-th column of layer $i+1$. Similarly, if the $j$-th column of layer $i$ is empty, the $j$-th row of layer $i-1$ and the corresponding entry in bias $i-1$ can be pruned. The same procedure can be applied to convolutional layers by removing the empty channels. 

\paragraph{Compressed Sparse Row Encoding} The weight matrices can be encoded using more compact representation uses 3 lists to store the non zero entries ($A$), the cumulative sum of number of non-zero entries for each row ($IR$) and the column indexes of the non-zero entries ($IC$). Note that since the matrix are sparse, storing the offset between the columns with a fixed amount of bits ($IC'$) is more efficient than storing the index. See appendix \ref{app:compression} for more details.

\paragraph{Huffman Encoding} Since the non-zero entries can assume at most $K-1$ different values, we can create an optimal encoding by using an Huffman coding scheme. This decreased the expected number of bits used to store the entries of $A'$ from 32 (or 64 depending on the floating point number representation) to $k \le \log K+1$.

\section{Experiments}
In preliminary experiments, we compare the compression ratio obtained with the procedure described in the previous section and th accuracy achieved by different Bayesian compression techniques on the well studied MNIST dataset with dense (LeNet300-100) and convolutional (LeNet5) architectures. The details of the training procedure are described in appendix \ref{app:training}.

% Starting from a pre-trained model trained with L2 norm, the models have been re-trained for 200 epochs using the Sparse Variational Dropout procedure. 
% The Soft Weight Sharing re-training methodology involved the use of 17 mixture components, one of which has been fixed to zero with a fixed mixing proportion $\pi_0 = 0.999$. A gamma hyper-prior ($\alpha=10^5$, $\beta=10$) have been applied to the precision of the Gaussian components to ensure numerical stability and the KL-Divergence term has been scaled down by a coefficient $\tau = 10^{-2}$. The Soft Weight Sharing models have been trained for 100 epochs.
% Further details regarding the optimization procedure and computation of the compression ratio can be found in the appendices.

% %We ...... [EXPERIMENT specifications] %MARCO%%%%%%%

The results presented in table \ref{fig:res} suggests that the joint procedure presented in this paper results in a dramatic increase of the compression ratio, achieving state-of-the-art results for both the dense and convolutional network architectures. Further work could address the interaction between the soft-weight sharing methodology and structured sparsity inducing techniques \citep{WenWWCL16,louizos2017bayesian} to reduce the overhead introduced by the compression format.
\begin{table}[!ht]
%   \begin{varwidth}[b]{0.6\linewidth}
    \centering
    \caption{Test accuracy (Acc), percentage of non-zero weights and compression ratio (CR) evaluated on the MNIST classification task. \textit{Soft-weight sharing} (SWS) \citep{2017arXiv170204008U} and the \textit{Variational Dropout} (VD) \citep{2017arXiv170105369M} approaches are compared with the Ridge regression (L2) and the combined approach proposed in this work (VD+SWS). All the accuracies are evaluated on the compressed models. The figure reports the best accuracies and compression ratios obtained by the three different techniques on the LeNet5 architecture by changing the values of the hyper-parameters.\\}
\label{tab:res}
%     \begin{tabular}{lllll}
% \hline
% Architecture                  & Training & Acc $[\%]$& $\frac{|W\neq0|}{|W|}$ $[\%]$ & CR  \\ \hline
% \multirow{4}{*}{LeNet300-100} & L2              & \bf{98.39}    & 100                    & 1   \\ 
%                               & SWS             & 98.16    & 8.6                    & 34  \\ 
%                               & VD              & 98.05    & 1.6                   & 131   \\ 
%                               & VD+SWS          & 98.24    & \bf{1.5}                    & \bf{161} \\ \hline
% \multirow{4}{*}{LeNet5}       & L2              & 99.13    & 100                    & 1   \\ 
%                               & SWS             & 99.01    & 3.6                    & 84  \\ 
%                               & VD              & 99.09    & 0.7                      & 349  \\  
%                               & VD+SWS          & \bf{99.14}    & \bf{0.5}                    & \bf{482} \\ \hline
% \end{tabular}
%     \label{table:res}
%   \end{varwidth}%
%   \hfill
%   \begin{minipage}[b]{0.29\linewidth}
%     \centering
%     \captionof{figure}{Effect on accuracy and compression ratio of the choice of the hyper-parameters for the LeNet5 architecture.\\}
%     \includegraphics[width=1.15\textwidth]{imgs/a_vs_cr.png}
%     \label{fig:res}
%   \end{minipage}
\begin{tabular}{llllll}
\cline{1-5}
Architecture                  & Training & Acc $[\%]$   & $\frac{|W\neq0|}{|W|}$ $[\%]$ & CR         & \multirow{9}{*}{\raisebox{-22.51mm}[0pt][0pt]{\includegraphics[width=0.323\textwidth]{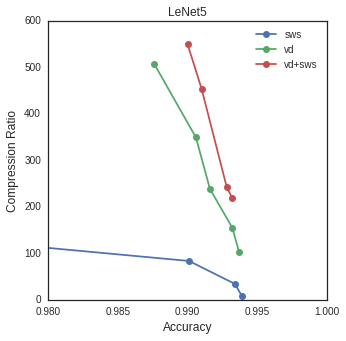}}
     \label{fig:res}} \\ \cline{1-5}
\multirow{4}{*}{LeNet300-100} & L2       & \bf{98.39} & 100                           & 1          &                    \\
                              & SWS      & 98.16        & 8.6                           & 34         &                    \\
                              & VD       & 98.05        & 1.6                           & 131        &                    \\
                              & VD+SWS   & 98.24        & \bf{1.5}                    & \bf{161} &                    \\ \cline{1-5}
\multirow{4}{*}{LeNet5}       & L2       & 99.13        & 100                           & 1          &                    \\
                              & SWS      & 99.01        & 3.6                           & 84         &                    \\
                              & VD       & 99.09        & 0.7                           & 349        &                    \\
                              & VD+SWS   & \bf{99.14} & \bf{0.5}                    & \bf{482} &                    \\ \cline{1-5}
\end{tabular}
\end{table}

\bibliography{bibliography}
\begin{appendices}
\section{Training}
\label{app:training}
The full training objective equation is given by the variational lower bound in equation \ref{eq:elbo}, where the two KL-Divergence terms have been scaled according to two coefficients $\tau_1$ and $\tau_2$ respectively:
\begin{align*}
\mathcal{L}\lp \phivec,\psivec \rp &=L_{\mathcal{D}}\lp\phi\rp  -\sum_{i=1}^D\lp \tau_1 D_{KL}\lp \mathcal{N}\lp w_i|\theta_i,\sigma_i^2\rp||\frac{1}{\lvert w_i\rvert}\rp +\tau_2\log GM_{\psivec}\lp m_i=\theta_i \rp\rp+C
\end{align*}
Starting from a pre-trained model, in a first warm-up phase we set $\tau_1=1$ and $\tau_2 = 0$. Note that this part of the training procedure matches the Sparse Variational Dropout methodology \citep{2017arXiv170105369M}. 
After reaching convergence (200 epochs in our experiments), we initialize the parameters $\psivec$ for the mixture model and the coefficient $\tau_2$ is set to a value of $2\ 10^{-2}$ ($\tau_1$ is kept to $1$) to induce the clustering effect with the Soft Weight sharing procedure.
This phase usually requires 50-100 epochs to reach convergence.

The mixture model used for our experiments uses 17 components, one of which has been fixed to zero with a fixed mixing proportion $\pi_0 = 0.999$. A gamma hyper-prior ($\alpha=10^5$, $\beta=10$)  have been applied to the precision of the Gaussian components to ensure numerical stability.

The proposed parametrization stores the weight variance $\boldsymbol{\sigma}^2$, each mixture component precision $\boldsymbol{\lambda}$ and the mixing proportions $\boldsymbol{\pi}$ in the logarithmic space.
All the models have been trained using ADAM optimizer, the learning rates and initialization values for the parameters are reported in  Table~\ref{tab:lrs}

\begin{table}[!h]
\centering
\caption{Learning rates and initialization values corresponding to the model parameters. $\wvec$ represents the weight vector of a pre-trained model, while $\Delta\mu$ represent the distance between the means of the mixing components and it is obtained by dividing two times the standard deviation of the $\wvec$ distribution by the number of mixing components $K$. Note that the indexing $i$ goes from $1$ to $D$ while $k$ starts from $-\frac{K-1}{2}$ and reaches $\frac{K-1}{2}$.\\}
\label{tab:lrs}
\begin{tabular}{llllll}
&$\theta_i$     & $\log\sigma_i^2$ & $\mu_k$     & $\log\lambda_k$ & $\log\pi_k$        \\ \hline
Initialization & $w_i$ & -10            & $k\ \Delta\mu $& $-2\log\lp0.9\  \Delta\mu\rp$ & $\begin{cases}\log0.999\ \ \ k=0\\\log\frac{1-\pi_0}{K}\ \ \ k\neq0\end{cases}$\\\\
Learning Rate  & $5\ 10^{-5}$      & $10^{-4}$      & $10^{-4}$                                                   & $10^{-4}$                                     & $3\ 10^{-3}$                                      
\end{tabular}
\end{table}
\newpage{}

\section{Details on the Compression Scheme}
\label{app:compression}
%TODO MENTION
% If the offset exceeds the number of bits used for the representation, a $0$ entry can be added to $A$ 
\subsection{Clustering}
The first part of the compression pipeline slightly differs for the model trained without a mixture of Gaussian prior (VD in table \ref{tab:res}). As suggested in \citet{2017arXiv170105369M}, once the model reaches convergence, all the parameters with a binary dropout rate $b_i=\sigma_i^2/\lp\theta_i^2+\sigma_i^2\rp\ge t$ are set to 0 ($t=0.95$ in our experiments).
Furthermore, since this methodology does not enforce clustering, in order to fairly compare the final compression ratio, the $\thetavec$ values are clustered with a 64 mixture components (no training for $\thetavec$ is involved) and quantized accordingly. The increased number of clusters has been selected to not compromise the performances of the discretized network. The other steps of the pipeline do not differ for the 3 evaluated methodologies.
\subsection{Compressed Sparse Row Format}
As mentioned in section \ref{sec:compression}, the Compressed Sparse Row format used in this work stores the column indexes by representing the offset with a fixed amount of bits (5 bits for the LeNet300-100 architecture, while 8 bits for the convolutional LeNet5 version).
If the distance between two non-zero entries exceeds the pre-defined number of bits, a zero entry is inserted in $A$. Note that because of this procedure, the total number of codewords in the Huffman codebook is $K$.
No optimization has been applied to the index array $IC$ since it didn't result into significant improvement. For further details regarding the optimized CSR format, refer to \citet{2015arXiv151000149H} and \citet{2017arXiv170204008U}.

\section{Derivation of the KL-Divergence}

%Context
In order to compute the KL-divergence between the joint prior and approximate posterior distribution, we use the factorization reported in table \ref{tab:distr}:
\label{app:kl}
\begin{align*}
-D_{KL}\lp q_{\boldsymbol{\phivec}}\lp\wvec,\mvec\rp || p_{\psivec}\lp \wvec,\mvec\rp\rp &= \int\int q_{\phivec}\lp \wvec,\mvec\rp\log \frac{p_{\psivec}\lp \wvec,\mvec\rp}{q_{\phivec}\lp \wvec,\mvec\rp}d\wvec\ d\mvec\\
&=\sum_{i=1}^D \int\int  q_{\sigma_i}\lp w_i|m_i\rp\ q_{\theta_i}\lp m_i\rp \log \frac{p\lp w_i\rp p_{\psivec}\lp m_i\rp}{ q_{\sigma_i}\lp w_i|m_i\rp\ q_{\theta_i}\lp m_i\rp }dm_i\ dw_i\\
&=\sum_{i=1}^D\int\lp\int  q_{\sigma_i}\lp w_i|m_i\rp  q_{\theta_i}\lp m_i\rp \log \frac{p\lp w_i\rp}{ q_{\sigma_i}\lp w_i|m_i\rp} dm_i\rp dw_i \\&\ \ +\sum_{i=1}^D\int\lp\int  q_{\sigma_i}\lp w_i|m_i\rp\ dw_i\rp q_{\theta_i}\lp m_i\rp \log \frac{p_{\psivec}\lp m_i\rp}{\ q_{\theta_i}\lp m_i\rp } dm_i\\
\end{align*}
By plugging in $q_{\theta_i}\lp m_i\rp = \delta_{\theta_i}\lp m_i\rp$ and observing $\int  q_{\sigma_i}\lp w_i|m_i\rp\ dw_i = 1$, we obtain:
\begin{align}
-D_{KL}\lp q_{\boldsymbol{\phivec}}\lp\wvec,\mvec\rp || p_{\psivec}\lp \wvec,\mvec\rp\rp &=\sum_{i=1}^D\int q_{\sigma_i}\lp w_i|m_i=\theta_i\rp \log \frac{p\lp w_i\rp}{ q_{\sigma_i}\lp w_i|m_i=\theta_i\rp} dw_i\nonumber\\&\ \ \ +\sum_{i=1}^D\int\ q_{\theta_i}\lp m_i\rp \log \frac{p_{\psi}\lp m_i\rp}{\ q_{\theta_i}\lp m_i\rp } dm_i\nonumber\\
&=-\sum_{i=1}^D\lp D_{KL}\lp q_{\sigma_i}\lp w_i|m_i=\theta_i\rp||p\lp w_i\rp\rp +D_{KL}\lp q_{\theta_i}\lp m_i\rp || p_{\psivec}\lp m_i\rp\rp\rp
\label{eq:kl}
\end{align}
Where the first KL-divergence can be approximated according to \citet{2017arXiv170105369M}:
\begin{align}
D_{KL}\lp q_{\sigma_i}\lp w_i|m_i=\theta_i\rp||p\lp w_i\rp\rp  &= D_{KL}\lp \mathcal{N}\lp w_i|\theta_i,\sigma_i^2\rp||\frac{1}{\lvert w_i\rvert}\rp + C \nonumber\\
&\approx k_1 \sigma\lp k_2 + k_3 \log\frac{\sigma_i^2}{\theta_i^2}\rp-0.5\log\lp1+\frac{\theta_i^2}{\sigma^2}\rp+C
\label{eq:kl1}
\end{align}
\begin{align*}
\text{With}\ \ \ \ \ k_1 = 0.63576,\ \ \ \ k_2=1.87320,\ \ \ \ k_3=1.48695
\end{align*}
While the second term can be computed by decomposing the KL-divergence into the entropy of the approximate posterior $\mathcal{H}\lp q_{\theta_i}\lp m_i\rp\rp$ and the cross-entropy between prior and posterior distributions $\mathcal{H}\lp q_{\theta_i}\lp m_i\rp, p_{\psivec}\lp m_i\rp\rp$.
Note that the entropy of a delta distribution does not depend on the parameter $\theta_i$, therefore it can be considered to be constant.
\begin{align}
D_{KL}\lp q_{\theta_i}\lp m_i\rp || p_{\psivec}\lp m_i\rp\rp &= -\mathcal{H}\lp q_{\theta_i}\lp m_i\rp\rp + \mathcal{H}\lp q_{\theta_i}\lp m_i\rp, p_{\psivec}\lp m_i\rp\rp\nonumber\\
&=\underbrace{-\mathcal{H}\lp \delta_{\theta_i}\lp m_i\rp\rp}_{C} - \int \log p_{\psivec}\lp m_i\rp  \delta_{\theta_i}\lp m_i\rp dm_i\nonumber\\
&= - \log p_{\psivec}\lp m_i=\theta_i\rp + C \nonumber\\
&= \log GM\lp m_i=\theta_i|\psivec\rp + C\nonumber\\
&= \log\sum_{k=1}^K \pi_k\ \mathcal{N}\lp m_i=\theta_i | \mu_k, \lambda_k^{-1}\rp + C
\label{eq:kl2}
\end{align}
Plugging-in the results from equations \ref{eq:kl1} and \ref{eq:kl2} into equation \ref{eq:kl} we obtain the full expression for the KL-divergence reported in equation \ref{eq:klend}.

\end{appendices}

\end{document}